\documentclass[10pt,twocolumn,letterpaper]{article}

\usepackage{cvpr}              

\usepackage{graphicx}
\usepackage{amsmath}
\usepackage{amssymb}
\usepackage{booktabs}
\usepackage{nicefrac}
\usepackage{float}
\usepackage[accsupp]{axessibility}  

\usepackage[table]{xcolor}
\definecolor{blue}{HTML}{0055cc}
\definecolor{red}{HTML}{cc1100}   
\definecolor{orange}{HTML}{cc7700}
\usepackage[pagebackref,breaklinks, colorlinks,linkcolor=red,citecolor=blue]{hyperref}
\usepackage[capitalize]{cleveref}
\crefname{section}{Sec.}{Secs.}
\Crefname{section}{Section}{Sections}
\Crefname{table}{Table}{Tables}
\crefname{table}{Tab.}{Tabs.}
\def\cvprPaperID{52} 
\def\confName{CVPR}
\def\confYear{2023}

\usepackage{times}
\usepackage{epsfig}
\usepackage{graphicx}
\usepackage{wrapfig}
\usepackage{amsmath,amssymb,amsthm}
\usepackage{algorithm,algorithmicx,algpseudocode}
\usepackage{bm,xspace}
\usepackage{comment}
\usepackage{verbatim}
\usepackage{multirow}
\usepackage{balance}
\usepackage{url}
\usepackage{booktabs}
\usepackage{etoolbox,siunitx}
\usepackage{calc}
\usepackage{pifont,hologo}
\usepackage{color}

\newcommand{\figref}[1]{Fig.~\ref{#1}}

\newcommand{\secref}[1]{Sec.~\ref{#1}}

\usepackage[super]{nth}
\usepackage{nicefrac}
\robustify\bfseries

\def\mM{\mathcal{M}}

\def\pascal{$\text{PASCAL-}5^i$}
\def\coco{$\text{COCO-}20^i$}
\def\wideline{0.6pt}

\DeclareMathSymbol{@}{\mathord}{letters}{"3B}

\def\latex/{\LaTeX}
\def\bibtex/{\hologo{BibTeX}}

\makeatletter
\DeclareRobustCommand\onedot{\futurelet\@let@token\@onedot}
\def\@onedot{\ifx\@let@token.\else.\null\fi\xspace}
 
\def\ie{\emph{i.e}\onedot}

\newcommand*{\Rom}[1]{\expandafter\@slowromancap\romannumeral #1@}
\newcommand*{\rom}[1]{\expandafter\romannumeral #1}

\def\1{\bm{1}}

\def\mC{{\bm{C}}}
\def\mD{{\bm{D}}}

\def\mF{{\bm{F}}}

\def\mI{{\bm{I}}}

\def\mK{{\bm{K}}}

\def\mM{{\bm{M}}}

\def\mQ{{\bm{Q}}}

\def\mS{{\bm{S}}}

\def\mV{{\bm{V}}}
\def\mW{{\bm{W}}}
\def\mX{{\bm{X}}}

\DeclareMathAlphabet{\mathsfit}{\encodingdefault}{\sfdefault}{m}{sl}
\SetMathAlphabet{\mathsfit}{bold}{\encodingdefault}{\sfdefault}{bx}{n}

\newcommand{\R}{\mathbb{R}}

\begin{document}
 
\title{Hierarchical Dense Correlation Distillation for Few-Shot Segmentation \\
Extended Abstract}

\author{
Bohao Peng\textsuperscript{1}, Zhuotao Tian\textsuperscript{2}\thanks{Corresponding Author},
Xiaoyang Wu\textsuperscript{3}, Chenyao Wang\textsuperscript{1}, Shu Liu\textsuperscript{2},
Jingyong Su\textsuperscript{4}, Jiaya Jia\textsuperscript{1,2} \\ \\ 
\textsuperscript{1}The Chinese University of Hong Kong\quad 
\textsuperscript{2}SmartMore \\
\textsuperscript{3}The University of Hong Kong \quad
\textsuperscript{4}Harbin Institute of Technology, Shenzhen \\
}
 
\maketitle 

\begin{abstract}
Few-shot semantic segmentation (FSS) aims to form class-agnostic models segmenting unseen classes with only a handful of annotations. Previous methods limited to the semantic feature and prototype representation suffer from coarse segmentation granularity and train-set overfitting. In this work, we design Hierarchically Decoupled Matching Network (HDMNet) mining pixel-level support correlation based on the transformer architecture. The self-attention modules are used to assist in establishing hierarchical dense features, as a means to accomplish the cascade matching between query and support features. Moreover, we propose a matching module to reduce train-set overfitting and introduce correlation distillation leveraging semantic correspondence from coarse resolution to boost fine-grained segmentation. Our method performs decently in experiments. We achieve $50.0\%$ mIoU on \coco~dataset one-shot setting and $56.0\%$ on five-shot segmentation, respectively. The code will be available on the project website. We hope our work can benefit broader industrial applications where novel classes with limited annotations are required to be decently identified.
\end{abstract} 

\section{Introduction}
\label{sec:intro}

 
Semantic segmentation tasks~\cite{jiang2021semi,cui2022generalized,lai2021cac,apd,lai2022decouplenet,tian2023cac,cui2022region,zhang2022mediseg} have made tremendous progress in recent years, benefiting from the rapid development of deep learning. However, most existing deep networks are not scalable to previously unseen classes and rely on annotated datasets to achieve satisfying performance.

\begin{figure}
    \centering
    \includegraphics[width=0.4\textwidth]{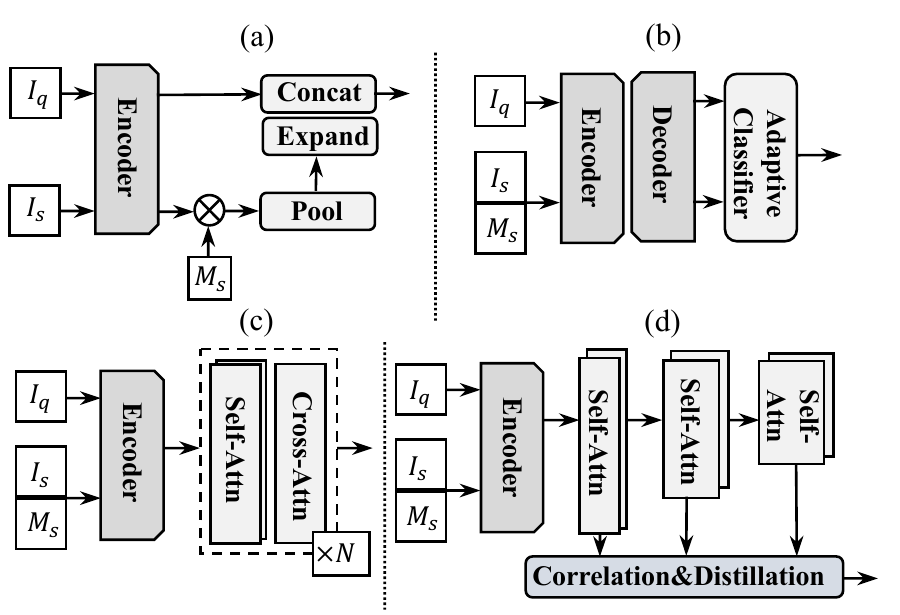}
    \caption{Illustration of different few-shot segmentation frameworks. (a) Prototype-based method. (b) Adaptive-classifier method. (c) Feature matching with transformer architecture. (d) The proposed HDMNet with correlation map distillation.}
   \vspace{-0.3in}
    \label{fig:architecture_comparison}
\end{figure}

Previous few-shot learning methods may still suffer from coarse segmentation granularity and train-set overfitting~\cite{tian2020prior} issue that stems from framework design, as illustrated in \figref{fig:architecture_comparison} (a)-(b).
Concretely, prototype-based~\cite{wang2019panet,tian2022generalized,luo2021pfenet++,zhang2019canet} and adaptive-classifier methods~\cite{boudiaf2021RePRI} aim at distinguishing different categories with global class-wise characteristics. It is challenging to compute the correspondence of different components between query and support objects for the dense prediction tasks. In contrast, matching-based methods~\cite{zhang2021cycle} mine pixel-level correlation but may heavily rely on class-specific features and cause overfitting and weak generalization.

To address these issues, we propose Hierarchically Decoupled Matching Network (HDMNet) with correlation map distillation for better mining pixel-level support correspondences. HDMNet extends transformer architecture~\cite{vaswani2017attention} to construct the feature pyramid and performs dense matching. Previous transformer-based methods~\cite{zhang2021cycle} adopt the self-attention layer to parse features and then feed query and support features to the cross-attention layer for pattern matching, as illustrated in~\figref{fig:architecture_comparison}(c). This process stacks the self- and cross-attention layers multiple times, mixes separated embedding features, and accidentally causes unnecessary information interference. 
Therefore, we instead decouple the feature parsing and matching process in a hierarchical paradigm and design a new matching module based on correlation and distillation that can alleviate the train-set overfitting problem. 
 

 \begin{figure*}[!ht]
    \centering
    \includegraphics[width=0.8\textwidth,height=0.3\textwidth]{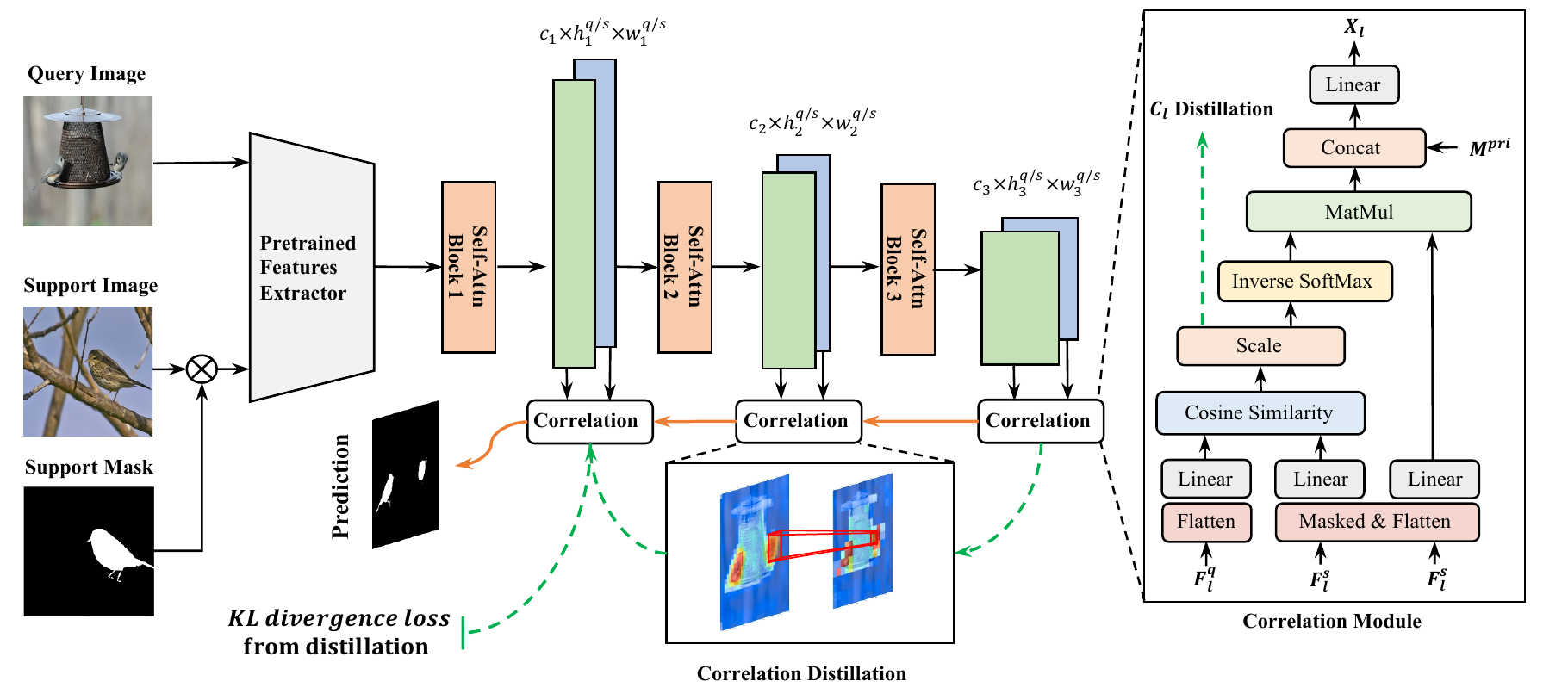}
    \vspace{-0.2in}
    \caption{Schematic overview of the proposed few-shot semantic segmentation model. $h_l^{\nicefrac{q}{s}}=\frac{H^{\nicefrac{q}{s}}}{2^{l+2}}$, $w_l^{\nicefrac{q}{s}}=\frac{W^{\nicefrac{q}{s}}}{2^{l+2}}$ indicate the height and width of the $l$-th stage features, and $H^{\nicefrac{q}{s}}$ and $W^{\nicefrac{q}{s}}$ are the height and width of the input query and support images. $c_l$ represents the channels and $c_{l+1}>c_{l}$.}
    \label{fig:architecture}
   \vspace{-0.2in}
\end{figure*}

\section{Preliminary}
\label{sec:task}

Few-shot segmentation is to train segmentation for novel objects with only a few annotated support images. In definition, the model is trained on ${\mD_{train}}$ and is evaluated on $\mD_{test}$. Suppose the category sets in $\mD_{train}$ and  $\mD_{test}$ are $\mC_{train}$ and $\mC_{test}$ respectively. There is no intersection between the training and testing sets, \ie, $\mC_{train} \cap \mC_{test} = \emptyset$. Following previous work, episodes are applied to both train set $\mD_{train}$ and test set $\mD_{test}$. 
  
Each episode is composed of a query set $\mQ = \{(\mI^q, \mM^q)\}$ and a support set $\mS = \{(\mI^s_i, \mM^s_i)\}_{i=1}^K$ with the same class $c$ , where $\mI^q,\mI^s \in \mathbb{R}^{H\times W\times 3} $ represent the RGB images and $\mM^q,\mM^s \in \mathbb{R}^{H\times W}$ denote their binary masks. Both the query masks $\mM^q$ and the support masks $\mM^s$ are used during the training process, while only the support masks $\mM^s$ are accessible in testing. Since the model parameters are fixed and require no optimization for novel categories during testing, the model is trained to leverage the semantic clues provided by the support set to locate the regions of interest on the query images. 

\section{Our Method}
In this section, we present the details of our proposed Hierarchically Decoupled Matching Network (HDMNet), which consists of novel yet effective designs of the decoupled matching structure and a correlation distillation. 

\subsection{Hierarchically Decoupled Matching Network}
\label{sec:method}

Given the query set $\mQ=\{(\mI^q, \mM^q)\}$ and the support set $\mS=\{(\mI^s_i, \mM^s_i)\}_{i=1}^K$, HDMNet adopts a parameter-fixed encoder to extract rich features of the query and support images, following~\cite{tian2020prior}. The difference is on the design of new decoder to yield predictions on the query images by decently leveraging pixel-level feature matching between the query and support sets. The pipeline is shown in~\figref{fig:architecture}. 

\subsection{Overview of the Architecture}
\label{sec:MSSMStructure}

\paragraph{Motivation.}

\begin{figure}
    \centering
    \includegraphics[width=0.4\textwidth]{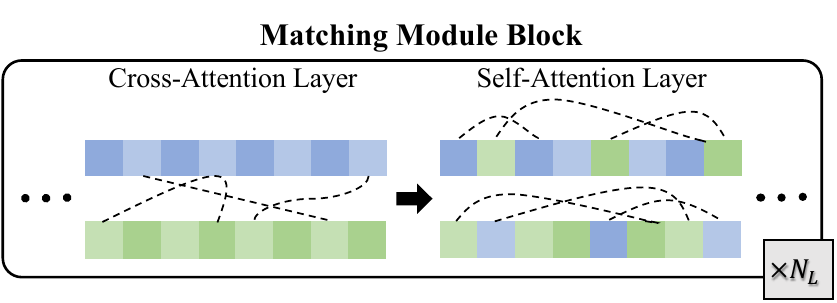}
    \caption{Feature Matching with directly stacking cross-attention layers and self-attention layers. It's intuitive to notice that the cross-attention layer mixes the query and support features, destroying the purity of parsing and matching consistency.}
    \label{fig:directly_stack}
    \vspace{-0.3cm}
\end{figure}

The stacked self- and cross-attention layers used by previous methods may lead to inferior results in that the necessary support information may be accumulated to the distracters, making the decoder harder to distinguish among them as shown in Fig.~\ref{fig:directly_stack}. 
To ensure the purity of the sequential features and consistency of pattern matching, we propose a new hierarchically matching structure decoupling the down-sampling and matching processes, where only independent self-layers are adopted to build hierarchical features.  


\vspace{-0.1in}
\paragraph{Decoupled downsampling and matching.}
First, the query and support features extracted from the backbone are independently sent to sequential transformer blocks with only the self-attention layers to fully exploit self-correlation within the support and query features. We note that the down-sampling layers are inserted between blocks to establish a hierarchical structure that may assist in mining the inter-scale correlations.

Then, the intermediate feature maps of $L$ stages are collected, i.e., $\{\mF^q_l\}_{l=1}^L$ and $\{\mF^s_l\}_{l=1}^L$. Assume $\{\mF^q_l\}$ and $\{\mF^s_l\}$ have the same spatial size $[c_l\times h_l^{\nicefrac{q}{s}}\times w_l^{\nicefrac{q}{s}}]$ for simplicity's sake. $$h_l^{\nicefrac{q}{s}}=\frac{H^{\nicefrac{q}{s}}}{2^{l+2}},\quad w_l^{\nicefrac{q}{s}}=\frac{W^{\nicefrac{q}{s}}}{2^{l+2}},$$ $l$ is the stage index, and $c_l$ denotes the feature channel number. 
Finally, $\{\mF^q_l\}_{l=1}^L$ and $\{\mF^s_l\}_{l=1}^L$ are used to yield correlations $\{\mC_l\in \R^{h_l^qw_l^q\times h_l^sw_l^s}\}_{l=1}^L$ and enriched query features $\{\mX_l\in\R^{c_l\times h_l^q\times w_l^q}\}_{l=1}^L$. Detailed formulations are elaborated later in Eqs. \eqref{eq:corr_map} and \eqref{eq:corr_mechanism} in Sec.~\ref{sec:corr-mechanism}.

\vspace{-0.1in}
\paragraph{Coarse-grained to fine-grained decoder.} HDMNet incorporates a simple decoder to predict the final mask for the query image with the hierarchically enriched features $\{\mX_l\in\R^{c_l\times h_l^q\times w_l^q}\}_{l=1}^L$ in a coarse-to-fine manner. Specifically, the coarse-grained features $\mX'_{l+1}$ are scaled up to have the same spatial size as the fine-grained one $\mX'_{l}$. Then an MLP layer is adopted to fuse them with a residual connection, written as
\begin{equation}
    \mX'_l=\text{ReLU}(\text{MLP}(\mX_l+\zeta_l(\mX'_{l+1})))+\zeta_l(\mX'_{l+1}),  
    \label{eq:decoder} 
\end{equation}
where $l$ indicates the hierarchical stage, and $\zeta_l:\mathbb{R}^{H\times W}\mapsto\mathbb{R}^{h_l\times w_l}$ denotes the bilinear-interpolation resize function fitting the input size to that of the output. Finally, we apply a convolution layer with $1\times 1$ kernel size to $\mX'_1$ followed by a bilinear up-sampling layer to predict the query mask $\mM^{out}\in\mathbb{R}^{H\times W}$.

\subsection{Matching Module} 
\label{sec:corr-mechanism}

\paragraph{Attention.}
\label{sec:attention_function}
Following the general form~\cite{vaswani2017attention}, the critical element of the transformer block is  the dot-product attention layer, formulated as 
\begin{equation}
    \text{Attn}(\mQ,\mK,\mV) = \text{softmax}(\frac{\mQ\mK^T}{\sqrt{d}})\mV,
    \label{eq:attention_formulation}
\end{equation}
where $[\mQ;\mK;\mV]=[\mW^q\mF^q;\mW^k\mF^{s};\mW^v\mF^{s}]$, in which $\mF^q$ and $\mF^{s}$ denote the query and support features respectively, $\mW^q,\mW^k,\mW^v\in\mathbb{R}^{d\times d}$ are the learnable parameters, $d$ is the hidden dimension. 

The cross-attention layer takes essential support information from $\mV$, conditioned on the query-support correlation between $\mQ$ and $\mK$. When $\mF_q=\mF_{s}$, it functions as a self-attention layer for relating different positions within either the support or query input features.

\vspace{-0.1in}
\paragraph{Our correlation mechanism.}
\label{sec:correlation_mechanism}  
Our designed matching module based on the correlation mechanism retrieves the most relevant regions with high cosine similarity and fuses the high-level prior mask generated as that in~\cite{tian2020prior}. Given the query features $F^q$ and the support features $F^s$, we first transform the input features by
\begin{equation}
\begin{aligned}
    \hat{\mF}^q &= \varphi(\mF^q),\\
    \hat{\mF}^s &= \varphi(\mF^s\odot\mM^s),
\end{aligned}
\end{equation}
where $\odot$ is Hadamard product, $\varphi:\mathbb{R}^{c\times h\times w}\mapsto\mathbb{R}^{hw\times c}$ refers to the reshape function, and $\mM^s$ denotes the support mask. To mitigate the risk of overfitting the category-specific information brought by the feature norms, we measure the cosine similarities of the inner product angle, instead of performing dot product, to calculate the correlation map as $C\in\mathbb{R}^{h^qw^q\times h^sw^s}$ as
\begin{equation}
    \mC=\frac{\langle \mW^q\hat{\mF}^q,\mW^k\hat{\mF}^s\rangle }{\left\lVert \mW^q\hat{\mF}^q\right\rVert \left\lVert \mW^k\hat{\mF}^s\right\rVert t},
    \label{eq:corr_map}
\end{equation}
where $\mW^q,\mW^k\in\mathbb{R}^{c\times c}$ denote the learnable parameters, $\left\lVert\cdot\right\rVert$ indicates $L^2$ norm, and $t$ is a hyperparameter to control the distribution range, empirically set to 0.1 in all experiments. Inspired by~\cite{rocco2018neighbourhood,smith2017offline}, we propose the inverse softmax layer that normalizes the correlation along the query axis since we only retrieve the interested region of the query set:
\begin{equation}
    \hat{\mC}(i,j)=\frac{\text{exp}(\mC(i,j))}{\sum_{k=1}^{h_l^qw_l^q}\text{exp}(\mC(k, j))}.
\end{equation}

Finally, we introduce the prior mask $\mM^{pri}\in\mathbb{R}^{h^q\times w^q}$ calculated the same as~\cite{tian2020prior} by concatenating it with correspondence scores along the channel dimension to generate the matching results of
\begin{equation}
    \mX=\mW^o([\psi(\hat{\mC}(\mW^v\hat{\mF^s})),\mM^{pri}]),
    \label{eq:corr_mechanism} 
\end{equation}
where $\mW^v\in\mathbb{R}^{c\times c}$, $\mW^o\in\mathbb{R}^{c\times (c+1)}$ denote the learnable parameters, $\hat{\mF}^s\in\mathbb{R}^{h^sw^s\times c},\mX\in\mathbb{R}^{c\times h^q\times w^q}$ are flattened support features and matching output, and $\psi:\mathbb{R}^{h^qw^q\times c}\mapsto \mathbb{R}^{c\times h^q\times w^q}$ is the reshape function. 

\subsection{Correlation Map Distillation} 
 We propose correlation map distillation to encourage the correlation maps of earlier stages to retain the fine-grained segmentation quality without deprecating the contextual hints by facilitating the shallower ones to capture broader context information.

\vspace{-0.1in}
\paragraph{Distillation formulation.} 
\label{sec:correlation_distillation}
Eq.\eqref{eq:corr_map} calculates the correlation maps $\{\mC_l\in \R^{h_l^qw_l^q\times h_l^sw_l^s}\}_{l=1}^L$ for the query and support features. We reorganize $\mC_l$ with mean average and filter the irrelevant information by the support mask $M^s$ as
\begin{equation}
    \mC'_{l}(i) = \frac{\sum_{j=1}^{h_l^sw_l^s}\mC_{l}(i,j)\cdot[\varphi\circ\zeta_l(\mM^s)(j)>0]}{\sum_{j=1}^{h_l^sw_l^s}[\varphi\circ\zeta_l(\mM^s)(j)>0]},
    \label{eq:map_reorgain}
\end{equation}
where $l$ indicates the stage, $\mC'_l\in\R^{h_l^qw^q_l}$, and $\zeta_l$ is the resize function. Given flattened correlation maps, we apply a softmax layer to perform spatial normalization among all positions as
\begin{equation}
    \hat{\mC}'_l(i)=\frac{\text{exp}({\mC}'_l(i)/T)}{\sum_{j=1}^{h_l^qw_l^q}\text{exp}({\mC}'_l(j)/T)},
    \label{eq:softmax_normalization}
\end{equation}
where $l$ indicates the stage and $T$ denotes the temperature of distillation~\cite{hinton2015distilling} set to 1. Moreover, the results regarding the temperature $T$ are shown in the supplementary file.

Then the KL (Kullback-Leibler) divergence loss is used as supervision from the teacher to student with their softmax output. The correlation maps of adjacent stages act as the teacher and student respectively, formulated as
\begin{equation}
\begin{split}    
    \mathcal{L}_{KL} &= \sum_{x\in\mathcal{X}}\phi_{t}(x)\text{log}(\frac{\phi_{t}(x)}{\phi_{s}(x)}) \\
    &= \sum\nolimits_{i=1}^{h_l^qw_l^q}\zeta_l(\hat{\mC}_{l+1})(i)\cdot\text{log}(\frac{\zeta_l(\hat{\mC}_{l+1})(i)}{\hat{\mC}_{l}(i)}),
    \label{eq:kldiv_loss}
\end{split}
\end{equation}
where $l$ indicates the stage, $\phi_t$ is the teacher model while $\phi_s$ is the student model, and $\zeta_l:\mathbb{R}^{h_{l+1}^qw_{l+1}^q}\mapsto \mathbb{R}^{h_l^qw_l^q}$ represents resizing. In particular, the last correlation map without successor uses the ground truth as its teacher.

\begin{table}[!t]
    \small
    \centering
    \begin{tabular}{l l|c|c}
    \specialrule{\wideline}{0pt}{0pt}
    \hline
         \multirow{1}{*}{Backbone}&\multicolumn{1}{l|}{\multirow{1}{*}{Methods}}&\multicolumn{1}{c|}{1-shot}&\multicolumn{1}{c}{5-shot}\\
         \hline
         \multirow{3}{*}{VGG-16}
        &PFENet~\cite{tian2020prior}&36.3&40.4\\
         &BAM$^{\dag}$~\cite{lang2022BAM}&42.1&47.3\\
         &\textbf{HDMNet (Ours)}&\textbf{45.9}&\textbf{52.4}\\
    \hline
         \multirow{5}{*}{ResNet-50}
        &PFENet~\cite{tian2020prior}&35.8&39.0\\
        &RePRI~\cite{boudiaf2021RePRI}&34.1&41.6\\
         &BAM$^{\dag}$~\cite{lang2022BAM}&45.2&48.5\\
         &\textbf{HDMNet (Ours)}&\textbf{50.0}&\textbf{56.0}\\
    \specialrule{\wideline}{0pt}{0pt}
    \hline
    \end{tabular}
    \caption{Few-shot semantic segmentation performance comparison on COCO-$20^{i}$~\cite{nguyen2019featureweight}. $\dag$: Reproduced with $10,000$ test episodes for fair comparison.}
    \label{tab:coco_results}
    \vspace{-0.3cm}
\end{table}

\section{Experiments}
\label{sec:exp}

\begin{table}[!t]
    \small
    \centering
    \begin{tabular}{l l|c|c}
    \specialrule{\wideline}{0pt}{0pt}
    \hline
         \multirow{1}{*}{Backbone}&\multicolumn{1}{l|}{\multirow{1}{*}{Methods}}&\multicolumn{1}{c|}{1-shot}&\multicolumn{1}{c}{5-shot}\\
         \hline
         \multirow{3}{*}{VGG-16}
        &PFENet~\cite{tian2020prior}&58.0&59.0\\
         &HSNet~\cite{min2021hsnet}&59.7&64.1\\
         &\textbf{HDMNet (Ours)}&\textbf{65.1}&\textbf{69.3}\\
    \hline
         \multirow{5}{*}{ResNet-50}
         &HSNet~\cite{min2021hsnet}&64.0&69.5\\
         &PFENet~\cite{tian2020prior}&60.8&61.9\\
         &BAM~\cite{lang2022BAM}&67.8&70.9\\
         &\textbf{HDMNet (Ours)}&\textbf{69.4}&\textbf{71.8}\\
    \specialrule{\wideline}{0pt}{0pt}
    \hline
    \end{tabular}
    \vspace{-0mm}
    \caption{Performance on \pascal~\cite{shaban2017oneshot}.}
    \label{tab:pascal_results}
    \vspace{-2mm}
\end{table}

 \begin{table}[!t]
     \small
     \centering
     \scalebox{1.}{
     \begin{tabular}{cccccc}
     \specialrule{\wideline}{0pt}{0pt}
     \hline
          Ens.&HDM&Corr.&Distill&mIoU (\%)&$\Delta$ \\
          \hline
          &&&&44.7&0.0\\
          $\checkmark$&&&&45.8&+1.1\\
          $\checkmark$&$\checkmark$&&&47.9&+3.2\\
          $\checkmark$&$\checkmark$&$\checkmark$&&48.3&+3.6\\
          $\checkmark$&$\checkmark$&$\checkmark$&$\checkmark$&\textbf{50.0}&+\textbf{5.3}\\
     \specialrule{\wideline}{0pt}{0pt}
     \hline
     \end{tabular}
     }
     \caption{Ablation studies for HDMNet. The first line denotes the baseline.  Ens. means the ensemble strategy in BAM~\cite{lang2022BAM}. HDM means the framework mentioned in Sec. \ref{sec:MSSMStructure}; Corr. means the correlation mechanism  mentioned in~\secref{sec:correlation_mechanism}; Distill means the correlation distillation described in~\secref{sec:correlation_distillation}. }
     \label{tab:ablation_implemetation}
     \vspace{-0.1in}
 \end{table}

\begin{table}[!t]
     \small
     \centering
     \scalebox{.93}{
     \begin{tabular}{c|cccc}
     \specialrule{\wideline}{0pt}{0pt}
     \hline
     Decoder&mIoU(\%)&{params(\textit{M})}&time(\textit{ms})&FLOPs(\textit{G})\\
        \hline
        CyCTR~\cite{zhang2021cycle}&40.3&5.6&54.3&96.7\\
        HSNet~\cite{min2021hsnet}&39.2&2.6&25.5&20.6\\
        BAM~\cite{lang2022BAM}&45.2&4.1&\bf{7.4}&26.0\\
        \hline
        Ours-$S_1$&47.1&\bf{1.3}&15.1&\bf{8.8}\\
        Ours-$S_2$&48.8&2.1&21.1&10.2\\
        Ours-$S_3$&\bf{50.0}&2.8&27.4&10.6\\
        Ours-$S_4$&48.6&3.6&38.0&10.6\\
     \specialrule{\wideline}{0pt}{0pt}
     \hline
     \end{tabular}
     }
     \caption{Comparison of decoders from different methods in terms of accuracy, efficiency, and model size. $S_i$ indicates constructing our decoder with $i$ matching stages.}
     \label{tab:ablation_levels}
 \end{table}

\paragraph{Comparison with state-of-the-Art methods. }
Following the setting of~\cite{tian2020prior}, we use two benchmark few-shot segmentation datasets, i.e., $\text{PASCAL-}5^i$~\cite{shaban2017oneshot} and $\text{COCO-}20^i$~\cite{nguyen2019featureweight}, to evaluate HDMNet. Implementation details can be found in our full paper.
In Tables \ref{tab:coco_results} and \ref{tab:pascal_results}, HDMNet achieves competitive results against the classic, \ie, PFENet (TPAMI 2020) and cutting-edge, \ie, BAM (CVPR 2022) methods.

\vspace{-0.3cm}
\paragraph{Ablation study.} Table~\ref{tab:ablation_implemetation} shows ablation results regarding the effectiveness of different components and architecture design, where the mIoU results are averaged over four splits. All ablation experiments are conducted under~\coco~ 1-shot with ResNet-50. 

\begin{figure}
     \centering
     \includegraphics[width=0.4\textwidth]{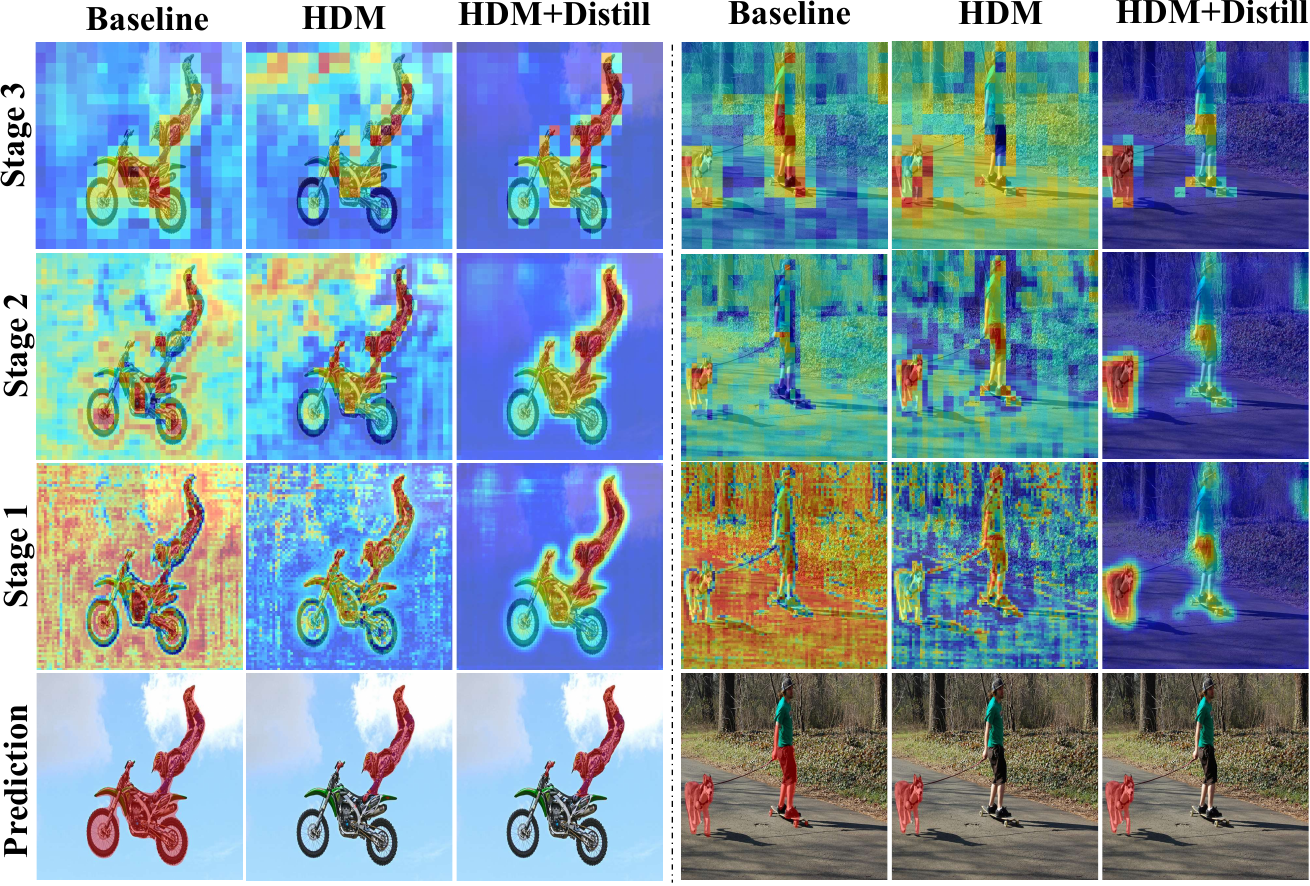}
     \caption{Qualitative correlation maps in 1-3 matching stages. The target classes of the left and right panels are ``people'' and ``dog''.}
     \label{fig:ablation_pyramid_level}
     \vspace{-0.3cm}
 \end{figure}

Table~\ref{tab:ablation_levels} compares the decoders of previous methods and our proposed matching pyramid with different stage numbers in terms of accuracy, efficiency, and model size. 
~\figref{fig:ablation_pyramid_level} visualizes qualitative results of correlation maps in 1-3 matching stages under distinct designs compared in Table~\ref{tab:ablation_levels}.

\section{Conclusion}
\label{sec:con}
HDMNet decouples the downsampling and matching process to prevent information interference. Further, we designed a novel matching module constructed on the correlation mechanism and distillation. Experiments demonstrate that our designs alleviate the training-class overfitting problem and improve the generality to unseen classes.


{\small
\bibliographystyle{ieee_fullname}
\bibliography{egbib_simple}
}

\end{document}